\pdfoutput=1
\documentclass[11pt]{article}

\usepackage{acl}

\usepackage{times}
\usepackage{latexsym}
\usepackage{booktabs}
\usepackage{xcolor}
\usepackage{multirow}
\usepackage[T1]{fontenc}
\usepackage[utf8]{inputenc}
\usepackage{microtype}
\usepackage{graphicx}
\usepackage{soul}
\usepackage{amsmath}
\usepackage{amsfonts}
\usepackage{array}
\usepackage{algorithm}
\usepackage{algorithmic}
\usepackage{wasysym}
\usepackage{multirow}
\usepackage{enumitem}
\usepackage[switch]{lineno}
\usepackage{tabularx}
\usepackage[normalem]{ulem}
\usepackage[para]{footmisc}
\usepackage[english]{babel}
\author{Thai Le \\
  Penn State University \\
  \texttt{thaile@psu.edu} \\\And
  Noseong Park \\
  Yonsei University \\
  \texttt{noseong@yonsei.ac.kr} \\\And
    Dongwon Lee \\
  Penn State University \\
  \texttt{dongwon@psu.edu} \\}
  
\title{SHIELD: Defending Textual Neural Networks against Multiple Black-Box Adversarial Attacks with Stochastic Multi-Expert Patcher}
\newcolumntype{H}{>{\setbox0=\hbox\bgroup}c<{\egroup}@{}}
\newcolumntype{s}{>{\hsize=.1\hsize}X}

\newcommand{\mymethod}{\textsc{\textsf{Shield}}}

\definecolor{byellow}{rgb}{0.92, 0.7, 0}

\begin{document}
\maketitle
\begin{abstract}
Even though several methods have proposed to defend textual neural network (NN) models against black-box adversarial attacks, they often defend against a specific text perturbation strategy and/or require re-training the models from scratch. This leads to a lack of generalization in practice and redundant computation. In particular, the state-of-the-art transformer models (e.g., BERT, RoBERTa) require great time and computation resources. By borrowing an idea from \textit{software engineering}, in order to address these limitations, we propose a novel algorithm, {\mymethod}, which modifies and re-trains only the last layer of a textual NN, and thus it ``patches'' and ``transforms'' the NN into a stochastic weighted ensemble of multi-expert prediction heads. Considering that most of current black-box attacks rely on iterative search mechanisms to optimize their adversarial perturbations, {\mymethod} confuses the attackers by automatically utilizing different weighted ensembles of predictors depending on the input. In other words, {\mymethod} breaks a fundamental assumption of the attack, which is a victim NN model remains constant during an attack. By conducting comprehensive experiments, we demonstrate that all of CNN, RNN, BERT, and RoBERTa-based textual NNs, once patched by {\mymethod}, exhibit a relative enhancement of 15\%--70\% in accuracy on average against 14 different black-box attacks, outperforming 6 defensive baselines across 3 public datasets. All codes are to be released. 
% \lee{if allowed by ACL, put them into anonymous github and put the URL here}
\end{abstract}

% Most of previous works (e.g., \cite{le2021sweet,zhou2021defense,keller2021bert,pruthi2019combating,dong2021towards,mozes2020frequency,Wang2021AdversarialTW} in adversarial defense are designed either for a specific type (e.g., word, synonym substitution, misspellings) or level (e.g., character or word-based) of attack. Thus, they are usually evaluated against a small subset of (${\leq}4$) attack methods. Even though there are works that propose general defense methods, they are often built upon \textit{adversarial training}~\cite{goodfellow2014explaining} which requires training everything from scratch (e.g., \cite{sibetter,miyato2016adversarial,zhang2017mixup} or limited to a set of predefined attacks (e.g., \cite{zhou2019learning}). Even though, adversarial training-based defense works well against different attacks on BERT and RoBERTa, its performance is far out-weighted by {\mymethod} (Table \ref{tab:results_short}). 

\section{Introduction}
\noindent \textbf{Adversarial Text Attack and Defense.} 
After being trained to maximize prediction performance, textual NN models frequently become vulnerable to adversarial attacks \cite{papernot2016limitations,wang2019towards}. In the NLP domain, in general, adversaries utilize different strategies to perturb an input sentence such that its semantic meaning is preserved while successfully letting a target NN model output a desired prediction. Text perturbations are typically generated by replacing or inserting critical words (e.g., HotFlip~\cite{ebrahimi2017hotflip}, TextFooler~\cite{jin2019bert}), characters (e.g., DeepWordBug \cite{gao2018black}, TextBugger~\cite{li2018textbugger}) in a sentence or by manipulating a whole sentence (e.g., SCPNA~\cite{iyyer2018adversarial}, GAN-based\cite{zhengli2018iclr}). 
% Different from white-box attacks such as HotFlip~\cite{ebrahimi2017hotflip} where parameters of a target NN model are accessible, black-box attacks assume no access to such parameters and hence are more practical. 

Since many recent NLP models are known to be vulnerable to adversarial black-box attacks (e.g., fake news detection \cite{malcom,zhou2019fake}, dialog systems \cite{cheng2019evaluating}, and so on), robust defenses for textual NN models are required. Even though several papers have proposed to defend NNs against such attacks, they were designed for either a specific type of attack (e.g., word or synonym substitution~\cite{Wang2021AdversarialTW,dong2021towards,mozes2020frequency,zhou2021defense}, misspellings~\cite{pruthi2019combating}, character-level~\cite{pruthi2019combating}, or word-based~\cite{le2021sweet}). Even though there exist some general defensive methods, most of them enrich NN models by re-training them with adversarial data augmented via known attack strategies ~\cite{miyato2016adversarial,liu2020robust,pang2019mixup} or with external information such as knowledge graphs \cite{li2019knowledge}.

However, these augmentations often induce substantial overhead in training or are still limited to only a small set of predefined attacks (e.g., \cite{zhou2019learning}). Hence, we are in search of defense algorithms that directly enhance NN models' structures (e.g., \cite{li2019knowledge}) while achieving higher generalization capability without the need of acquiring additional data. 

\begin{figure*}[t!]
\centerline{\includegraphics[width=1.0\textwidth]{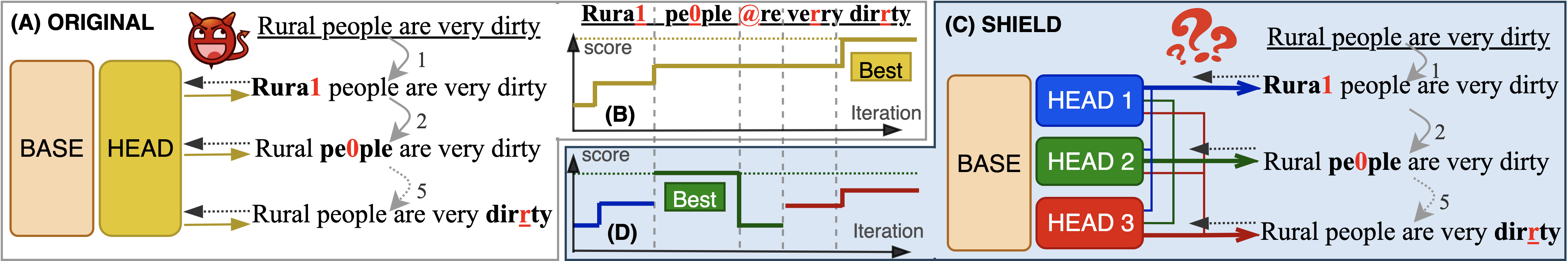}}
\caption{Motivation of {\mymethod}: An attacker optimizes a step objective function (score) to search for the best perturbation by iteratively replacing each of the original 5 tokens with a perturbed one. \textbf{(A)} The attacker assumes the model \emph{remains unchanged} and \textbf{(B)} gives coherent signal during the iteration search, resulting in the true \textit{best} attack: ``dirty''$\rightarrow$``dir\textbf{\textcolor{red}{r}}ty''. \textbf{(C)} A model patched with {\mymethod} utilizes a weighted ensemble of 3 diverse heads depending on the input. Therefore, the ensemble weights keep changing over time during adversaries' perturbation search processes -- the line width represents the ensemble weights. \textbf{(D)} {\mymethod} confuses the attacker with 3 \textit{varying} distributions of the score, resulting in a sub-optimal attack ``people''$\rightarrow$``pe\textbf{\textcolor{red}{0}}ple''.}
\label{fig:motivation}
% \vspace{-5pt}
\end{figure*}

\noindent \textbf{Motivation (Fig. \ref{fig:motivation})}. Different from white-box attacks, black-box attacks do not have access to a target model's parameters, which are crucial for achieving effective attacks. Hence, attackers often query the target model repeatedly to acquire the necessary information for optimizing their strategy. From our analyses of 14 black-box attacks published during 2018--2020 (Table \ref{tab:compare_attacks}), all of them, except for  \textit{SCPNA}~\cite{iyyer2018adversarial},  rely on a searching algorithm (e.g., greedy, genetic) to iteratively replace each character/word in a sentence with a perturbation candidate to optimize the choice of characters/words and how they should be crafted to attack the target model (Fig. \ref{fig:motivation}A). Even though this process is effective in terms of attack performance, they assume that the model's parameters remain ``unchanged'' and the model outputs 
%{\color{red}coherent}
``coherent''
signals during the iterative search (Fig. \ref{fig:motivation}A and \ref{fig:motivation}B). 
Our key intuition is, however, {\em to obfuscate the attackers by breaking this assumption}. Specifically, we want to develop an algorithm that automatically utilizes a diverse set of models during inference. This can be done by training multiple sub-models instead of a single prediction model and randomly select one of them during inference to obfuscate the iterative search mechanism. However, this then introduces impractical computational overhead during both training and inference, especially when one wants to maximize prediction accuracy by utilizing complex SOTA sub-models such as BERT \cite{devlin2018bert} and RoBERTa \cite{liu2019RoBERTa}. Moreover, it also does not guarantee that trained models are sufficiently diverse to fool attackers. Furthermore, applying this strategy to existing NN models would also require re-training everything from the scratch, 
%which may not be practical.
rendering the approach impractical.

\renewcommand{\tabcolsep}{1.5pt}
\begin{table}[t]
\centering
\small
\label{tab:notations}
\begin{tabular}{Hlcccc}
% \hline
\toprule
\multirow{2}{*}{\textbf{\#}} & \multirow{2}{*}{\textbf{Attack Method}} & {\textbf{Search}} & \textbf{Atk} & \textbf{Sem.} & \textbf{Natr.} \\
 & {} & \textbf{Method} & \textbf{Level} & \textbf{Presv.} & \textbf{Presv.} \\
\cmidrule(lr){1-6}
% {HotFlip}\cite{ebrahimi2017hotflip} & Gradient & Word/Char & White Box & $\checkmark$\\
% \textit{HotFlip$^*$} & Grads & Word & White Box & -\\
% {FGSM} & Gradient & Word & White Box & -\\
% 1 & Reduction (2018) \cite{feng2018pathologies} & Greedy & Word \\
1 & SCPNA~\citeauthor{iyyer2018adversarial} & {TP} & SN & $\checkmark$ & $\checkmark$\\
2 & TextBugger(TB)~\citeauthor{li2018textbugger} & GD & CR & $\checkmark$ & \\
3 & DeepWordBug(DW)~\citeauthor{gao2018black} & GD & CR & $\checkmark$ &  \\
4 & Kuleshov~\citeauthor{kuleshov2018adversarial} & GD & WD & $\checkmark$ & $\checkmark$ \\
5 & TextFooler(TF)~\citeauthor{jin2019bert}  & GD & WD & $\checkmark$ & \\
6 & IGA~\citeauthor{wang2019natural} & {GN} & WD & & \\
7 & Pruthi~\citeauthor{pruthi2019combating} & GD & CR\\
8 & PWWS(PS)~\citeauthor{ren2019generating} & GD & WD\\
9 & Alzantot~\citeauthor{alzantot2018generating} & {GN} & WD & & $\checkmark$\\
10 & BAE~\citeauthor{garg2020bae} & GD & WD & $\checkmark$\\
11 & BERT-Atk(BERTK)~\citeauthor{li2020bert} & GD & WD & $\checkmark$\\
12 & PSO~\citeauthor{zang2020word} & {GN} & WD \\
13 & Checklist~\citeauthor{ribeiro2020beyond} & GD & WD \\
14 & Clare~\citeauthor{li2020contextualized} & GD & WD & $\checkmark$ & $\checkmark$ \\ 
% \hline
\bottomrule
\multicolumn{4}{l}{TP: Template; GD: Greedy; GN: Genetics}\\
\multicolumn{4}{l}{CR: Character; WD: Word; SN: Sentence}
% \multicolumn{4}{l}{(-) SCPNA relies on a fixed set of templates }
% \multicolumn{4}{l}{*We use word-based attack for HF and Black Box attack for TB}
\end{tabular}
\caption{Different attack methods with i) how they search for adversarial perturbations, ii) their attack level, and iii) whether they maintain the original semantics (\textit{Sem. Presv.}), pursue the naturalness of the perturbed sentence (\textit{Natr. Presv.)}, or both of them.}
\label{tab:compare_attacks}
% \vspace{-5pt}
\end{table}

\noindent \textbf{Proposal.} To address these challenges, we borrow ideas from \textit{software engineering} where bugs can be readily removed by an external installation patch. Specifically, we develop a novel neural patching algorithm, named as {\mymethod}, which patches only the last layer of an already deployed textual NN model (e.g., CNN, RNN, transformers\cite{vaswani2017attention,bahdanau2014neural}) and transforms it into an ensemble of multi-experts or \textit{prediction heads} (Fig. \ref{fig:motivation}C). During inference, then {\mymethod} automatically utilizes a stochastic weighted ensemble of experts for prediction depending on inputs. This will obfuscate adversaries' perturbation search, 
%initiated by the adversaries, 
making black-box attacks much more difficult regardless of attack types, e.g., character or word level attacks (Fig. \ref{fig:motivation}C,D). By patching only the last layer of a model, {\mymethod} also introduces lightweight computational overhead and requires no additional training data. In summary, our contributions %in this paper 
are as follows:

\begin{itemize}[leftmargin=\dimexpr\parindent+0.1\labelwidth\relax]
\setlength\itemsep{0.5pt}
% \item We first formalize the concept of \textit{Neural Patching}, a novel mechanism to fix or enhance a \textit{trained} NN model $f$ by modifying only a part of $f$'s architecture or parameters (Figure \ref{fig:patching}).
\item We propose {\mymethod}, a novel neural patching algorithm that transforms a already-trained NN model
%$f$ 
to a \textit{stochastic} ensemble of multi-experts with little computational overhead. %compared to the previous ensemble methods in the literature. 
\item We demonstrate the effectiveness of {\mymethod}. CNN, RNN, BERT, and RoBERTa-based textual models patched by {\mymethod} achieve an increase of 15\%--70\% on their robustness across 14 different black-box attacks, outperforming 6 defensive baselines on 3 public NLP datasets.
\item To the best of our knowledge, this work by far includes the most comprehensive evaluation for the defense against black-box attacks.
\end{itemize}

\section{The Proposed Method: {\mymethod}}

% \begin{figure}[tb]
% \centerline{\includegraphics[width=0.32\textwidth]{architecture.png}}
% \caption{Architecture of {\mymethod} with $K$ experts.}
% \label{fig:architecture}
% \end{figure}

We introduce \textit{Stochastic Multi-Expert Neural Patcher} (\mymethod) which patches \textit{only} the last layer of an already \textit{trained} NN model $f(\mathbf{x}, \theta)$ and transforms it into an ensemble of multiple expert predictors with stochastic weights. These predictors are designed to be strategically selected with different weights during inference depending on the input. This is realized by two complementary modules, namely (i) a \textit{Stochastic Ensemble (SE)} module that transforms $f(\cdot)$ into a randomized ensemble of different heads and (ii) a \textit{Multi-Expert (ME)} module that uses Neural Architecture Search (NAS) to dynamically learn the optimal architecture of each head to promote their diversity. 
%We describe {\mymethod} in detail as follows.

\subsection{A Stochastic Ensemble (SE) Module}\label{sec:SE}
This module extends the last layer of $f(\cdot)$, which is typically a fully-connected layer (followed by a softmax for classification), to an ensemble of $K$ \textit{prediction heads}, denoted $\mathcal{H}{=}\{h(\cdot)\}_j^K$. Each head $h_j(\cdot)$, parameterized by $\theta_{h_j}$, is an expert predictor that is fed with a feature representation learned by \textit{up to the second-last layer} of $f(\cdot)$ and outputs a prediction logit score:
% \park{You cannot use $f$ here. $f$ means a full network but you mean here a network up to its second-last layer. So, you have to use something like $f'$ to distinguish from $f$. If not, will be very confusing.}
\begin{equation}
 h_j: f(\mathbf{x}, \theta^*_{L-1}) \in \mathbb{R}^{Q} \mapsto \tilde{y}_j \in \mathbb{R}^{M},
 \label{eqn:head}
\end{equation}
where $\theta^*_{L-1}$ are \textit{fixed} parameters of $f$ up to the last prediction head layer, $Q$ is the size of the feature representation of $\mathbf{x}$ generated by the base model $f(\mathbf{x}, \theta^*_{L-1})$, and $M$ is the number of labels. To aggregate all logit scores returned from all heads, then, a classical ensemble method would average them as the final prediction: $\hat{y}^*{=}\frac{1}{K} \sum_j^K \tilde{y}_j$. However, this simple aggregation assumes each $h_j(\cdot) \in \mathcal{H}$ learns from very similar training signals. Hence, when $\theta^*_{L-1}$ already learns some of the task-dependent information, $\mathcal{H}$ will eventually converge \textit{not} to a set of experts but very similar predictors. To resolve this issue, we introduce stochasticity into the process by assigning \textit{prediction heads with stochastic weights} during both training and inference. Specifically, we introduce a new aggregation mechanism:
\begin{equation}
\hat{y} = \frac{1}{K} \sum_j^K \alpha_j w_j \tilde{y}_j,
 \label{eqn:final_logit}
\end{equation}

where $w_j$ weights $\tilde{y}_j$ according to head $j$'s expertise on the current input $\mathbf{x}$, and $\alpha_j \in [0,1]$ is a probabilistic scalar, representing how much of the weight $w_j$ should be accounted for. Let us denote $w$, $\alpha \in \mathbb{R}^{K}$ as vectors containing all scalars $w_j$ and $\alpha_j$, respectively, and $\tilde{\mathbf{y}} \in \mathbb{R}^{(K\times M)}$ as the concatenation of all vectors $\tilde{y}_j$ returned from each of the heads. We calculate $w$ and $\alpha$ as follows:
\begin{equation}
 \begin{aligned}
w &= \mathbf{W}^T(\tilde{\mathbf{y}} \oplus f(\mathbf{x}, \theta^*_{L-1})) + \mathbf{b},
 \end{aligned}
 \label{eqn:scale_score}
\end{equation}
\vspace{-15pt}
\begin{equation}
 \alpha = \mathrm{softmax}((w + \mathbf{g})/\tau),
 \label{eqn:gumbel_softmax}
\end{equation}
\noindent where $\mathbf{W} \in \mathbb{R}^{(K\times M + Q)\times K}$, $\mathbf{b} \in \mathbb{R}^{K}$ are trainable parameters, $\mathbf{g} \in \mathbb{R}^K$ is a noise vector sampled from the \textit{Standard Gumbel Distribution} and therefore, probability vector $\alpha$ is sampled by a technique known as \textit{Gumbel-Softmax}~\cite{jang2016categorical} controlled by the noise vector $\mathbf{g}$ and the temperature $\tau$. Unlike the standard Softmax, the Gumbel-Softmax is able to learn a categorical distribution (over $K$ heads) optimized for a downstream task~\cite{jang2016categorical}. Annealing $\tau{\rightarrow}0$ encourages a pseudo one-hot vector (e.g., [0.94, 0.03, 0.01, 0.02] when $K{=}4$), which makes Eq. (\ref{eqn:final_logit}) a mixture of experts~\cite{avnimelech1999boosted}. Importantly, $\alpha$ is sampled in an inherently stochastic way depending on the gumbel noise $\mathbf{g}$.
% In contrast, as $\tau{\rightarrow}\infty$, $\alpha$ becomes a uniform distribution and Eq. (\ref{eqn:final_logit}) weights each sub-model equally. 

% Here $\mathbf{W}, \mathbf{b}, s, \alpha$ is introduced to assign on how much each head contributes to the final prediction with some controlled noise. Vector $\alpha$ is sampled by Eq. (\ref{eqn:gumbel_softmax}) with the modified logit $w + \mathbf{g}$ and the inverse-temperature $\tau$. 
While $\mathbf{W}, \mathbf{b}$ is learned to \textit{deterministically} assigns more weights $w$ to heads that are experts for each input $\mathbf{x}$ (Eq. (\ref{eqn:scale_score})), $\alpha$ introduces \textit{stochasticity} into the final logits. The multiplication of $\alpha_jw_j$ in Eq. (\ref{eqn:final_logit}) then enables us to use different sets of weighted ensemble models \textit{while still maintaining the ranking of the most important head}. Thus, this further diversifies the learning of each expert and confuse attackers when they iteratively try different inputs to find good adversarial perturbations. 

Finally, to train this module, we use Eq. (\ref{eqn:final_logit}) as the final prediction and train the whole module with \textit{Negative Log Likelihood (NLL)} loss following the objective:
\begin{equation}
 \min_{\mathbf{W}, \mathbf{b}, \{\theta_{h}\}^K_j} \mathcal{L_{\mathrm{SE}}} = -\frac{1}{N} \sum_i^N y_i log(\mathrm{softmax}(\hat{y}_i)).
 \label{eqn:mle}
\end{equation}

% In previous works, a similar approach was used in generative adversarial networks~\cite{goodfellow2014generative} to train multiple generators, each of which produces only a specific type of images, e.g., short-hair male by the first generator, long-hair female by the second generator, and so on~\cite{10.5555/3304415.3304540}. In our case, we aim at learning a mixture of expert heads in order to make the original NN more robust against adversarial attacks.

% \begin{figure}[tb]
% \centerline{\includegraphics[width=0.53\textwidth]{scaler_clickbait_sst_smaller.png}}
% \caption{Effects of Scaler on Prediction Logits of a Single Expert}
% \label{fig:scaler}
% \end{figure}

\begin{algorithm}[tb]
 \footnotesize
 \caption{\textbf{Training {\mymethod} Algorithm}.}
 \label{alg:training}
 \begin{algorithmic}[1]
\STATE \textbf{Input:} pre-trained neural network $f(\cdot)$
\STATE \textbf{Input:} $\mathcal{O}$, $K$, $\tau$, $\gamma$
\STATE \textit{Initialize} $\mathbf{W}, \mathbf{b}, \theta_\mathcal{O}, \{\beta\}_j^K$
\REPEAT
\STATE Freeze $\{\beta\}_j^K$ and optimize $\mathbf{W}, \mathbf{b}, \theta_\mathcal{O}$ via Eq. (\ref{eqn:mle}) in mini-batch from \textit{train} set.

\STATE Freeze $\mathbf{W}, \mathbf{b}, \theta_\mathcal{O}$ and optimize $\{\beta\}_j^K$ via Eq. (\ref{eqn:me}) with $\gamma$ multiplier in mini-batch from \textit{validation} set.
\UNTIL{convergence}
 \end{algorithmic}
%  \vspace{-5pt}
\end{algorithm}

\subsection{A Multi-Expert (ME) Module}\label{sec:ME}

While the \textit{SE} module facilitates stochastic weighted ensemble among heads, the \textit{ME} module searches for the optimal architecture for each head that maximizes the diversity in how they make predictions. To do this, we utilize the \textsc{DARTS} algorithm~\cite{liu2018darts} as follows. Let us denote $\mathcal{O}_j{=}\{o_j(\cdot)\}_t^T$ where %the set of 
$T$ is the number of possible architectures to be selected for $h_j \in \mathcal{H}$. We want to learn a one-hot encoded \textit{selection} vector $\beta_j \in \mathbb{R}^T$ that assigns $h_j(\cdot) \leftarrow o_{j,\mathrm{argmax}(\beta_j)}(\cdot)$ during prediction. Since $\mathrm{argmax}(\cdot)$ operation is not differentiable, during training, we relax the categorical assignment of the architecture for $h_j(\cdot) \in \mathcal{H}$ to a softmax over all possible networks in $\mathcal{O}_j$:
\begin{equation}
 h_j(\cdot) \longleftarrow \frac{1}{T} \sum_t^T \frac{\mathrm{exp}(\beta_j^t)}{\sum_t^T \mathrm{exp}(\beta_j^T)} o_{j,t}(\cdot).
 \label{eqn:head_function}
\end{equation}

\noindent However, the original DARTS algorithm \textit{only} optimizes prediction performance. In our case, we also want to promote the diversity among heads. To do this, we force each $h_j(\cdot)$ to specialize in different features of an input, i.e., in how it makes predictions. This can be achieved by \textit{maximizing} the difference among the gradients of the word-embedding $\mathbf{e}_i$ of input $\mathbf{x}_i$ w.r.t to the outputs of each $h_j(\cdot) \in \mathcal{H}$. Hence, given a fixed set of parameters $\theta_{\mathcal{O}}$ of all possible networks for every heads, we train all selection vectors $\{\beta\}_j^K$ by optimizing the objective:
\begin{equation}
\begin{aligned}
&\mathrm{minimize}_{\{\beta\}_j^K} \mathcal{L}_{\mathrm{experts}} = \\
&\sum_i^N \sum_{n<m}^K{\Big(}\mathrm{d}(\nabla_{\mathbf{e}_i}\mathcal{J}_n; \nabla_{\mathbf{e}_i}\mathcal{J}_m) - ||\nabla_{\mathbf{e}_i}\mathcal{J}_n{-}\nabla_{\mathbf{e}_i}\mathcal{J}_m ||_2^2 \Big),
\end{aligned}
\label{eqn:expert}
\end{equation}

\noindent where $\mathrm{d}(\cdot)$ is the cosine-similarity function, and $\mathcal{J}_j$ is the NLL loss as if we only use a single prediction head $h_j$. In this module, however, not only do we want to maximize the differences among gradients vectors, but also we want to ensure the selected architectures eventually converge to good prediction performance. Therefore, we train the whole \textit{ME} module with the following objective:
\begin{equation}
 \mathrm{minimize}_{\{\beta\}_j^K} \mathcal{L}_{ME} = \mathcal{L}_{SE} + \gamma \mathcal{L}_{\mathrm{experts}}.
 \label{eqn:me}
\end{equation}

\subsection{Overall Framework}

To combine the \textit{SE} and \textit{ME} modules, we replace Eq. (\ref{eqn:head_function}) into Eq. (\ref{eqn:head}) and optimize the overall objective:
\begin{equation}
\begin{aligned}
&\mathrm{minimize}_{\mathbf{\{\beta\}_j^K}} \mathcal{L}_{\mathrm{ME}}^{\mathrm{val}} + \gamma \mathcal{L}_{\mathrm{experts}}^{\mathrm{val}} \quad
\mathrm{s.t.} \\
&\mathbf{W}, \mathbf{b}, \theta_{\mathcal{O}} = \mathrm{minimize}_{\mathbf{W}, \mathbf{b}, \theta_{\mathcal{O}}} \mathcal{L}_{\mathrm{SE}}^{\mathrm{train}}.
\end{aligned}
\label{eqn:final}
\end{equation}

\noindent We employ an \textit{iterative} training strategy~\cite{liu2018darts} with the \textit{Adam} optimization algorithm~\cite{kingma2014adam} as in Alg. \ref{alg:training}. By alternately freezing and training $\mathbf{W}, \mathbf{b}$, $\theta_{\mathcal{O}}$ and $\{\beta\}_j^K$ using a training set $\mathcal{D}_{\mathrm{train}}$ and a validation set $\mathcal{D}_{\mathrm{val}}$, we want to (i) achieve high quality prediction performance through Eq. (\ref{eqn:mle}) and (ii) select the optimal architecture for each expert to maximize their specialization through Eq. (\ref{eqn:expert}). 

% ####################################
\section{Experimental Evaluation}

\renewcommand{\tabcolsep}{2pt}
 \begin{table}[t]
 \centering
 \footnotesize
 \begin{tabular}{lcccc}
    \toprule
    {} & \textbf{\#Class} & \textbf{\#Vocab} & \textbf{\#Example}\\
    \cmidrule(lr){1-4}
     MR~\cite{pang2005seeing} & 2 & 19K  & 11K \\
     CB~\cite{anand2017we} & 2  & 25K & 32K \\
     HS~\cite{hateoffensive} & 3 & 35K & 25K \\
    \bottomrule
     \end{tabular}
    \caption{Statistics of experimental datasets.}
     \label{tab:dataset}
    %  \vspace{-10pt}
\end{table}

\subsection{Set-up}\label{sec:implementation}\label{sec:setup}
\noindent \textbf{Datasets \& Metric.} Table \ref{tab:dataset} shows the statistics of all experimental datasets: Clickbait detection (CB) \cite{anand2017we}, Hate Speech detection (HS) \cite{hateoffensive} and Movie Reviews classification (MR) \cite{pang2005seeing}. We split each dataset into \textit{train}, \textit{validation} and \textit{test} set with the ratio of 8:1:1 whenever standard public splits are not available. To report prediction performance on clean examples, we use the \textit{weighted F1} score to take the distribution of prediction labels into consideration. To report the robustness, we report prediction \textit{accuracy} under adversarial attacks~\cite{morris2020textattack}, i.e., \# of failed attacks over total \# of examples. A failed attack is only counted when the attacker fails to perturb (i.e., fail to flip the label of a \textit{correctly predicted} clean example).

\vspace{0.1in}
\noindent \textbf{Defense Baselines.} \label{sec:baselines}
We want to defend four textual NN models (base models) of different architectures, namely RNN with GRU cells~\cite{chung2014empirical}, \textit{transformer}-based BERT \cite{devlin2018bert} and RoBERTa \cite{liu2019RoBERTa}. We compare {\mymethod} with the following six defensive baselines:

\begin{itemize}[leftmargin=\dimexpr\parindent-0.1\labelwidth\relax]
\setlength\itemsep{0.5pt}
 \item \textit{Ensemble (Ens.)} is the classical ensemble of 5 different \textit{base models}. We use the average of all NLL losses from the base models as the final training loss. 
 \item \textit{Diversity Training (DT)}~\cite{kariyappa2019improving} is a variant of the \textit{Ensemble} baseline where a regularization term is added to maximize the coherency of gradient vectors of the input text w.r.t each sub-model. DT diversifies the feature-level expertise among heads.
 \item \textit{Adaptive Diversity Promoting (ADP)}~\cite{pang2019improving} is a variant of \textit{Ensemble} baseline where a regularization term is added to maximize the diversity among non-maximal predictions of individual sub-models. ADP diversifies the class-level expertise among heads.
 \item \textit{Mixup Training (Mixup)}~\cite{zhang2017mixup,sibetter} trains a \textit{base model} with data constructed by linear interpolation of two random training samples. In this work, we use \textit{Mixup} to regularize a NN to adapt linear transformation in-between the continuous embeddings of training samples.
 \item \textit{Adversarial Training (AdvT)}~\cite{miyato2016adversarial} is a semi-supervised algorithm that optimizes the NLL loss on the original training samples plus adversarial inputs.
 \item \textit{Robust Word Recognizer (ScRNN)}~\cite{pruthi2019combating} detects and corrects potential adversarial perturbations or misspellings in a text before feeding it to the base model for prediction.
\end{itemize}
\noindent \uline{Note that due to the insufficient memory of GPU Titian Xp to simultaneously train several BERT and RoBERTa sub-models, we exclude \textit{Ensemble}, \textit{DT}, and \textit{ADP} baseline for them.}

\renewcommand{\tabcolsep}{3.5pt}
\begin{table}[t!b]
 \centering
 \footnotesize
 \begin{tabular}{lcccc}
\toprule
\multicolumn{1}{c}{\textbf{Model/Dataset}} & \textbf{MR} & \textbf{HS} & \textbf{CB} & \textbf{AVG}\\
\cmidrule(lr){1-5}
RNN    &  0.73 &  0.88 &  \underline{0.97} &  0.86 \\
+Ensemble  &  \textbf{0.80} &  \textbf{0.90} &  \underline{0.97} &  \textbf{0.89} \\
+DT        &  \textbf{0.80} &  0.86 &  \underline{0.97} &  0.88 \\
+ADP       &  \textbf{0.80} &  0.88 &  \underline{0.97} &  \underline{0.88} \\
+Mixup     &  0.77 &  0.87 &  \underline{0.97} &  0.87 \\
+AdvT &  0.76 &  \underline{0.89} &  \textbf{0.98} &  0.88 \\
+ScRNN     &  {0.79} &  0.85 &  0.96 &  0.87 \\
\textbf{+\mymethod}     &  0.78 &  0.86 &  \underline{0.97} &  0.87 (${\uparrow}1.3\%$) \\
\cmidrule(lr){1-5}
CNN   &  0.719 &  \textbf{0.900} &  0.966 &  0.862 \\
+Ens.  &  0.770 &  0.881 &  \underline{0.975} &  0.875 \\
+DT        &  0.767 &  0.890 &  0.972 &  0.876 \\
+ADP       &  0.764 &  0.885 &  \textbf{0.977} &  0.875 \\
+Mixup     &  0.711 &  0.867 &  0.965 &  0.848 \\
+AdvT &  \underline{0.772} &  0.884 &  \textbf{0.977} &  \underline{0.878} \\
+ScRNN     &  0.758 &  0.854 &  0.972 &  0.861 \\
\textbf{+\mymethod}     &  \textbf{0.787} &  \underline{0.893} &  0.974 &  \textbf{0.885} (${\uparrow}2.7\%$)\\
\cmidrule(lr){1-5}
BERT       &  0.84 &  \underline{0.90} &  \textbf{1.00} &  \underline{0.91} \\
+Mixup     &  0.81 &  0.89 &  0.99 &  0.90 \\
+AdvT &  \underline{0.85} &  \textbf{0.91} &  0.99 &  \textbf{0.92} \\
+ScRNN     &  0.83 &  0.90 &  0.99 &  0.91 \\
\textbf{+\mymethod}     &  \textbf{0.86} &  0.90 &  \underline{0.99} &  \underline{0.91} (%${\uparrow}
$0\%$) \\
\cmidrule(lr){1-5}
RoBERTa    &  \underline{0.88} &  0.89 &  \textbf{1.00} &  \underline{0.92} \\
+Mixup     &  \textbf{0.88} &  \textbf{0.91} &  0.99 &  \textbf{0.93} \\
+AdvT &  0.87 &  0.89 &  0.99 &  0.92 \\
+ScRNN     &  0.88 &  \underline{0.90} &  0.99 &  0.92 \\
\textbf{+\mymethod}     &  0.88 &  0.89 &  \underline{0.99} &  0.92 (%${\uparrow}
$0\%$) \\
\bottomrule
 \end{tabular}
 \caption{Prediction F1 on clean examples. On average, {\mymethod} is still able to maintain the original fidelity.}
 \label{tab:fidelity_full}
%  \vspace{-15pt}
\end{table}

\vspace{0.1in}
\noindent \textbf{Attacks.} We comprehensively evaluate {\mymethod} under 14 different black-box attacks (Table \ref{tab:compare_attacks}). 
% We focus on testing with black-box attacks as they do not require access to the target model's parameters and gradients, hence they are more practical. 
These attacks differ in their attack levels (e.g., character, word, sentence-based), optimization algorithms for searching adversarial perturbations (e.g., through fixed templates, greedy, genetic-based search). Apart from lexical constraints such as limiting \# or \% of words to manipulate in a sentence, ignoring stop-words, etc., many of them also preserve the semantic meanings of a generated adversarial text via constraining the \textit{l2 distance} between its representation vector and that of the original text produced by either Universal Sentence Encoder (USE)~\cite{cer2018universal} or GloVe embeddings~\cite{pennington2014glove}. Moreover, to ensure that the perturbed texts still look natural, a few of the attack methods employ an external pre-trained language model (e.g., BERT\cite{devlin2018bert}, L2W~\cite{holtzman2018learning}) to optimize the log-likelihood of the adversarial texts. Due to computational limit, we only compare {\mymethod} with other baselines in 3 representative attacks, namely \textit{TextFooler}~\cite{jin2019bert}, \textit{DeepWordBug}~\cite{gao2018black} and \textit{PWWS}~\cite{ren2019generating}. They are among the most effective attacks. To ensure fairness and reproducibility, we use the external \textit{TextAttack}~\cite{morris2020textattack}
% ~\footnote{\url{https://github.com/QData/TextAttack}}
and \textit{OpenAttack}~\cite{zeng2020openattack}.
% ~\footnote{\url{https://github.com/thunlp/OpenAttack}} 
framework for adversarial text generation and evaluation.

\renewcommand{\tabcolsep}{3.2pt}
\begin{table*}[t!]
\centering
\small
\begin{tabular}{lcccccccccccccccccc}
\toprule
\textbf{Dataset} & \multicolumn{6}{c}{\textbf{Movie Reviews}} & \multicolumn{6}{c}{\textbf{Hate Speech}} & \multicolumn{6}{c}{\textbf{Clickbait}}  \\
\cmidrule(lr){2-7}\cmidrule(lr){8-13}\cmidrule(lr){14-19}
  %\textbf{Model}  
  & \multicolumn{2}{c}{\textbf{RNN}} & \multicolumn{2}{c}{\textbf{BERT}} & \multicolumn{2}{c}{\textbf{\small{RoBERTa}}} &  \multicolumn{2}{c}{\textbf{RNN}} & \multicolumn{2}{c}{\textbf{BERT}} & \multicolumn{2}{c}{\textbf{\small{RoBERTa}}}& \multicolumn{2}{c}{\textbf{RNN}} & \multicolumn{2}{c}{\textbf{BERT}} & \multicolumn{2}{c}{\textbf{\small{RoBERTa}}} \\
%   \cmidrule(lr){2-3}\cmidrule(lr){4-5}\cmidrule(lr){6-7}\cmidrule(lr){8-9}\cmidrule(lr){10-11}\cmidrule(lr){12-13}\cmidrule(lr){14-15}\cmidrule(lr){16-17}\cmidrule(lr){18-19}\cmidrule(lr){20-21}\cmidrule(lr){22-23}\cmidrule(lr){24-25}
\textbf{Attack} & \textbf{Bef.} & \textbf{Aft.} & \textbf{Bef.} & \textbf{Aft.} & \textbf{Bef.} & \textbf{Aft.} & \textbf{Bef.} & \textbf{Aft.} & \textbf{Bef.} & \textbf{Aft.} & \textbf{Bef.} & \textbf{Aft.} & \textbf{Bef.} & \textbf{Aft.} & \textbf{Bef.} & \textbf{Aft.} & \textbf{Bef.} & \textbf{Aft.} \\ 
% \cmidrule(lr){1-25}
% \textbf{W/o Attack} & \\
\cmidrule(lr){1-19}
SCPNA& 0.37 &\textcolor{red}{0.32}&  0.27 &\textcolor{red}{0.24}& 0.27 &\textbf{0.28}&0.51 &\textbf{0.72}&  0.25 &\textbf{0.29}& 0.23 &\textbf{0.3}& 0.51 &\textcolor{red}{0.5}&  0.44 &\textbf{0.49}&  0.4 &\textbf{0.4}\\
TB & 0.2 &\textbf{0.32}&  0.28 &\textbf{0.37}& 0.28 &\textbf{0.5}& 0.35 &\textbf{0.61}&  0.48 &\textbf{0.59}& 0.51 &\textbf{0.6}& 0.79 &\textbf{0.86}&  0.87 &\textbf{0.93}& 0.89 &\textbf{0.94}\\
DW  & 0.2 &\textbf{0.44}&  0.27 &\textbf{0.42}& 0.16 &\textbf{0.55}& 0.27 &\textbf{0.47}&  0.27 &\textbf{0.55}& 0.41 &\textbf{0.55}& 0.67 &\textbf{0.9}&  0.58 &\textbf{0.95}& 0.68 &\textbf{0.96}\\
Kuleshov& 0.01 &\textbf{0.12}&  0.07 &\textbf{0.22}& 0.05 &\textbf{0.28}& 0.04 &\textbf{0.18}&  0.09 &\textbf{0.28}& 0.03 &\textbf{0.25}& 0.37 &\textbf{0.71}&  0.52 &\textbf{0.88}& 0.63 &\textbf{0.9}\\
TF & 0.03 &\textbf{0.18}&  0.08 &\textbf{0.26}& 0.05 &\textbf{0.39}& 0.08 &\textbf{0.24}&  0.25 &\textbf{0.42}& 0.12 &\textbf{0.37}& 0.31 &\textbf{0.78}&  0.44 &\textbf{0.92}&  0.5 &\textbf{0.93}\\
IGA  & 0.05 &\textbf{0.29}&  0.16 &\textbf{0.32}& 0.13 &\textbf{0.5}& 0.16 &\textbf{0.34}&  0.27 &\textbf{0.35}& 0.24 &\textbf{0.33}&  0.6 &\textbf{0.8}&  0.79 &\textbf{0.95}& 0.77 &\textbf{0.96}\\
Pruthi  &0.53 &\textbf{0.56}&  0.48 &\textbf{0.49}& 0.54 &\textbf{0.54}& 0.59 &\textbf{0.71}&  0.45 &\textbf{0.59}& 0.53 &\textbf{0.59}& 0.94 &\textcolor{red}{0.92}&  0.96 &\textcolor{red}{0.95}& 0.96 &\textbf{0.96}\\
PS & 0.09 &\textbf{0.3}&  0.14 &\textbf{0.35}& 0.15 &\textbf{0.45}& 0.3 &\textbf{0.54}&  0.32 &\textbf{0.43}& 0.32 &\textbf{0.44}& 0.46 &\textbf{0.85}&  0.64 &\textbf{0.94}& 0.66 &\textbf{0.94}\\
Alzantot&  0.21 &\textbf{0.36}&  0.42 &\textbf{0.47}& 0.46 &\textbf{0.64}& 0.27 &\textbf{0.54}&  0.51 &\textbf{0.57}& 0.56 &\textcolor{red}{0.55}&0.73 &\textbf{0.83}&  0.92 &\textbf{0.97}&  0.9 &\textbf{0.98}\\
BAE  & 0.44 &\textbf{0.54}&  0.38 &\textbf{0.46}& 0.43 &\textbf{0.57}& 0.6 &\textbf{0.72}&  0.38 &\textbf{0.52}& 0.43 &\textbf{0.51}& 0.83 &\textbf{0.92}&0.4 &\textbf{0.81}& 0.39 &\textbf{0.92}\\
BERTK & 0.01 &\textbf{0.18}&  0.04 &\textbf{0.17}& 0.03 &\textbf{0.23}& 0.1 &\textbf{0.21}&  0.36 &\textbf{0.48}& 0.22 &\textbf{0.36}& 0.18 &\textbf{0.65}&  0.25 &\textbf{0.86}& 0.41 &\textbf{0.86}\\
PSO  & 0.05 &\textbf{0.07}&  0.14 &\textcolor{red}{0.12}& 0.07 &\textbf{0.15}& 0.35 &\textbf{0.54}&  0.38 &\textbf{0.4}& 0.35 &\textbf{0.4}& 0.6 &\textbf{0.64}&  0.75 &\textbf{0.87}& 0.71 &\textbf{0.87}\\
Checklist  & 0.7 &\textbf{0.76}&  0.84 &\textbf{0.85}& 0.88 &\textbf{0.88}& 0.86 &\textcolor{red}{0.81}&  0.89 &\textbf{0.89}& 0.88 &\textbf{0.88}& 0.98 &\textbf{0.98}&  0.99 &\textbf{1.0}&  1.0 &\textbf{1.0}\\
Clare& 0.16 &\textbf{0.35}&  0.23 &\textbf{0.28}& 0.27 &\textbf{0.54}& 0.76 &\textcolor{red}{0.72}&  0.79 &\textcolor{red}{0.78}& 0.72 &\textbf{0.76}&  0.7 &\textbf{0.87}&  0.48 &\textbf{0.86}& 0.68 &\textbf{0.94}\\
\cmidrule(lr){1-19}
\textbf{Average}  &  0.27 &\textbf{0.36}& 0.27 &\textbf{0.46}& 0.37 &\textbf{0.52}&  0.41 &\textbf{0.51}&  0.4 &\textbf{0.49}& 0.65 &\textbf{0.75}& 0.62 &\textbf{0.8}&  0.65 &\textbf{0.88}& 0.68 &\textbf{0.9}\\
\textit{Relative $\uparrow$\%} &\multicolumn{2}{r}{$\uparrow$54.55\%}&\multicolumn{2}{r}{$\uparrow$33.33\%}&\multicolumn{2}{r}{$\uparrow$70.37\%}&\multicolumn{2}{r}{$\uparrow$40.54\%}&\multicolumn{2}{r}{$\uparrow$24.39\%}&\multicolumn{2}{r}{$\uparrow$22.5\%}&\multicolumn{2}{r}{$\uparrow$29.03\%}&\multicolumn{2}{r}{$\uparrow$35.38\%}&\multicolumn{2}{r}{$\uparrow$32.35\%}\\
\bottomrule
\multicolumn{19}{l}{\textbf{Bold}, \textcolor{red}{Red}: \textbf{no worse} and \textcolor{red}{decreased} results from the base models}\\
\end{tabular}
\caption{Accuracy under adversarial attacks before (Bef.) and after (Aft.) patched with {\mymethod}. \textit{The results of CNN-based models are presented in the Appendix.}}
\label{tab:results}
% \vspace{-10pt}
\end{table*}

\vspace{0.1in}
\noindent \textbf{Implementation.} We train {\mymethod} of 5 experts ($K{=}5$) with $\gamma{=}0.5$. For each expert, we set $\mathcal{O}_j$ to 3 ($T{=}3$) possible networks: FCN with 1, 2 and 3 hidden layer(s). For each dataset, we use \textit{grid-search} to search for the best $\tau$ value from $\{1.0, 0.1, 0.01, 0.001\}$ based on the averaged defense performance on the \textit{validation} set under TextFooler~\cite{jin2019bert} and DeepWordBug~\cite{gao2018black}. We use 10\% of the training set as a separate development set during training with early-stop to prevent overfitting. We report the performance of \textbf{the best single model} \textit{across all attacks} on the test set. The Appendix includes all details on all models' parameters and implementation. We will release the code of {\mymethod}.

\subsection{Results} \label{sec:fidelity}
% \ul{}

\vspace{0.1in}
\noindent \textbf{Fidelity} We first evaluate {\mymethod}'s prediction performance \textit{without} adversarial attacks. Table \ref{tab:fidelity_full} shows that all base models patched by {\mymethod} still maintain similar F1 scores on average across all datasets. Although {\mymethod} with RNN has a slightly decrease in fidelity on Hate Speech dataset, this is negligible compared to the adversarial robustness benefits that {\mymethod} will provide (More below). 
% Except for \textit{ScRNN}, which observes decreases in performance under clean datasets in many cases, other baselines also maintain more or less the same accuracy with the base models. This shows that misspelling-based defenses such as \textit{ScRNN} might accidentally correct unintentional errors in the input text and lead to undesired predictions.

\vspace{0.1in}
\noindent \textbf{Computational Complexity} 
% Table~\ref{tab:space_complexity} shows the computational complexity of all methods in terms of memory and running time. 
Regarding the space complexity, {\mymethod} can extend a NN into an ensemble model with a marginal increase of \# of parameters. Specifically, with $B$ denoting \# of parameters of the base model, {\mymethod} has a space complexity of $\mathcal{O}(B{+}KU)$ while both \textit{Ensemble}, \textit{DT} and \textit{ADP} have a complexity of $\mathcal{O}(KB)$ and $U{\ll}B$. In case of BERT with $K{=}5$, {\mymethod} only requires an additional 8.3\%. While traditional ensemble methods require as many as 4 times additional parameters. During training, {\mymethod} \textit{only} trains $\mathcal{O}(KU)$ parameters, while other defense methods, including ones using data augmentation, update all of them. Specifically, with $K{=}5$, {\mymethod} only trains 8\% of the parameters of the base model and 1.6\% of the parameters of other BERT-based ensemble baselines. During inference, {\mymethod} is also 3 times faster than ensemble-based \textit{DT} and \textit{ADP} on average.

\vspace{0.1in}
\noindent \textbf{Robustness}\label{sec:robustness}
Table \ref{tab:results} shows the performance of {\mymethod} compared to the base models. Overall, \uline{{\mymethod} consistently improves the robustness of base models in 154/168 (~92\%) cases across 14 adversarial attacks regardless of their attack strategies.} Particularly, all CNN, RNN, BERT and RoBERTa-based textual models that are patched by {\mymethod} witness relative improvements in the average prediction accuracy from 15\% to as much as 70\%. Especially in the case of detecting clickbait, {\mymethod} can recover up to 5\% margin within the performance on clean examples in many cases. This demonstrates that {\mymethod} provides a versatile neural patching mechanism that can quickly and effectively defends against black-box adversaries \textit{without} making any assumptions on the attack strategies. 

We then compare {\mymethod} with all defense baselines under TextFooler (TF), DeepWordBug (DW), and PWWS (PS) attacks. These attacks are selected as (i) they are among the strongest attacks and (ii) they provide foundation mechanisms upon which other attacks are built. Table \ref{tab:results_short} shows that {\mymethod} achieves the best robustness across all attacks and datasets. On average, {\mymethod} observes an absolute improvement from +9\% to +18\% in accuracy over the second-best defense algorithms (DT in case of RNN, and AdvT in case of BERT, RoBERTa). Moreover, {\mymethod} outperforms other ensemble-based baselines (DT, ADP), and can be applied on top of a pre-trained BERT or RoBERTa model with only around 8\% additional parameters. However, that \# would increase to 500\% ($K{\leftarrow}5)$ in the case of DT and ADP, requiring over half a billion \# of parameters.

% \vspace{-10pt}
\section{Discussion}\label{sec:discussion}

% \renewcommand{\tabcolsep}{5pt}
% \begin{table}[]
%     \centering
%     \small
%     \begin{tabular}{lr}
%     \toprule
%         \multicolumn{1}{c}{\textbf{Dataset}} & \multicolumn{1}{c}{\textbf{Contribution of Head \#1--\#5}} \\
%         \cmidrule(lr){1-2}
%         Movie Reviews &  9.2\%, 1.6\%, 49.2\%, 5.6\%, 34.3\%  \\
%         Hate Speech &  4.5\%, 45.2\%, 5.4\%, 32.9\%, 12.0\% \\
%         Clickbait & 40.6\%, 23.6\%, 0.2\%, 12.1\%, 23.5\% \\
%     \bottomrule
%     \end{tabular}
%     \caption{Example of contribution of different heads in the final predictions by a {\mymethod} model}
%     \label{tab:head_contribution}
% \end{table}

\renewcommand{\tabcolsep}{1.35pt}
\begin{table}[t!]
 \centering
 \small
 \begin{tabular}{lccccccccccc}
\toprule
\multicolumn{1}{c}{\textbf{Dataset}} & \multicolumn{3}{c}{\textbf{MR}} & \multicolumn{3}{c}{\textbf{HS}} & \multicolumn{3}{c}{\textbf{CB}} & \multicolumn{1}{c}{\multirow{2}{*}{\textbf{AVG}}}\\
\cmidrule(lr){2-4}\cmidrule(lr){5-7}\cmidrule(lr){8-10}
\multicolumn{1}{c}{\textbf{Attack}} &  \textbf{TF} & \textbf{DW} & \textbf{PS} &\textbf{TF} & \textbf{DW} & \textbf{PS} & \textbf{TF} & \textbf{DW} & \textbf{PS} \\

\midrule
RNN &  0.02 &   0.2 &  0.09 & 0.09 &  0.26 &  0.32 & 0.31 &  0.67 &  0.46 & 0.27\\
+Ens.  &  0.01 &0.16 &  0.06 &  0.08 &\textcolor{red}{0.12} &  0.29 &  0.32 &0.66 &  0.48 & \textcolor{red}{0.24}\\
+DT&  \underline{0.03} &\underline{0.24} &\underline{0.1} &  \textbf{0.32} & \underline{0.53} &  \underline{0.53} &  \underline{0.35} &0.66 &0.5 & \underline{0.36}\\
+ADP  &  0.02 &0.18 &  0.09 &  0.18 &0.27 &  0.35 &  0.33 &0.66 &  0.47 & 0.28\\
+Mixup&  0.01 &0.14 &  \textcolor{red}{0.04} &  0.07 &0.42 &  0.29 &  0.27 &0.64 &  0.44 &\textcolor{red}{0.26}\\
+AdvT &  0.01 &0.3 &  0.09 &  0.17 &\textcolor{red}{0.18} &  0.35 &  0.33 &\underline{0.69} &  \underline{0.51} & 0.29\\
+ScRNN&  \underline{0.03} &0.17 &  0.08 &  0.15 &\textcolor{red}{0.16} &  0.32 &  0.33 &0.68 &  0.47 & 0.27\\
\textbf{+\mymethod} &\textbf{0.18} &\textbf{0.44} &\textbf{0.3} &\underline{0.26} &\textbf{0.61} &  \textbf{0.54} &\textbf{0.78} & \textbf{0.9} &  \textbf{0.85} & \textbf{0.54}\\

\midrule
CNN & 0.01 &  0.13 &  0.06 & 0.03 &   0.1 &  0.14 & 0.45 &   0.7 &  0.57 & 0.24\\
+Ens.  &  0.02 &0.16 &  0.07 &  \textcolor{red}{0.08} &\textcolor{red}{0.2} &  0.26 &  0.72 &\textbf{0.87} &  0.78 & 0.35\\
+DT&  \underline{0.03} &0.16 &  0.07 &  \textcolor{red}{0.08} &\textcolor{red}{\underline{0.25}} &  \underline{0.28} &  \textbf{0.75} &\textbf{0.87} &\underline{0.8} & \underline{0.37}\\
+ADP  &\textcolor{red}{0.0} &0.11 &  \textcolor{red}{0.04} &  \textcolor{red}{0.08} &\textcolor{red}{0.19} &  0.21 &  \textcolor{red}{0.19} &0.67 &  0.44 & \textcolor{red}{0.21}\\
+Mixup&  \underline{0.03} &\textcolor{red}{0.18} &0.1 &  \textcolor{red}{0.07} &\textbf{0.32} &  0.24 &  \textcolor{red}{0.13} &0.6 &  \textcolor{red}{0.37} & \textcolor{red}{0.23}\\
+AdvT &  0.02 &0.17 &  0.07 &\underline{0.1} &0.18 &  0.27 &  0.33 &0.73 &  0.55  & 0.27\\
+ScRNN&  \underline{0.03} &\underline{0.24} &  \underline{0.11} &  \textcolor{red}{0.06} &\textcolor{red}{0.14} &  0.22 &  0.36 &0.69 &  0.54 & 0.27\\
\textbf{+\mymethod} & \textbf{0.19} &  \textbf{0.38} &  \textbf{0.28} & \textbf{0.19} &  \textbf{0.32} &  \textbf{0.34} & \underline{0.74} &  \underline{0.86} &  \textbf{0.81} & \textbf{0.46}\\

\midrule
BERT  &  0.09 &0.2 &  0.19 &  0.26 &0.16 &  0.38 &  0.49 &0.5 &  0.49 & 0.31\\
+Mixup&  \underline{0.11} &0.3 &  \underline{0.22} &  \textcolor{red}{0.15} &0.19 &  \textcolor{red}{0.22} &  \textcolor{red}{0.39} &\textcolor{red}{0.48} &  0.57 & \textcolor{red}{0.29}\\
+AdvT &  \underline{0.11} &\underline{0.25} &  0.19 &  \underline{0.37} &\underline{0.47} &  \textbf{0.47} &  \underline{0.69} &\underline{0.73} &  \underline{0.81} & \underline{0.45}\\
+ScRNN&  \textcolor{red}{0.03} &\textcolor{red}{0.11} &  \textcolor{red}{0.13} &  \textcolor{red}{0.34} &0.33 &  0.34 &  \textcolor{red}{0.41} &0.51 &0.6 & 0.31\\
\textbf{+\mymethod} & \textbf{0.26} &  \textbf{0.42} &  \textbf{0.35} & \textbf{0.42} &  \textbf{0.55} &  \underline{0.43} & \textbf{0.92} &  \textbf{0.95} &  \textbf{0.94} & \textbf{0.58}\\

\midrule
RoBERTa&  0.06 &0.18 &  0.16 &0.1 &0.12 &  0.12 &  0.37 &0.34 &  0.45 & 0.21\\
+Mixup&  \textcolor{red}{0.05} &\textcolor{red}{0.16} &  \textcolor{red}{0.15} &  0.17 &\underline{0.43} &  0.32 &  0.52 &0.69 &  0.66 & 0.35\\
+AdvT &\underline{0.1} &\underline{0.21} &  \underline{0.21} &  \underline{0.34} &\underline{0.43} &  \underline{0.42} &  \underline{0.67} &\underline{0.79} &  \underline{0.77} & \underline{0.44}\\
+ScRNN &  \textcolor{red}{0.04} &0.18 &  \textcolor{red}{0.15} &  0.19 &0.38 &  0.32 &  0.57 &0.74 &0.7 & 0.36\\
\textbf{+\mymethod}& \textbf{0.39} &  \textbf{0.55} &  \textbf{0.45} & \textbf{0.37} &  \textbf{0.55} &  \textbf{0.44} & \textbf{0.93} &  \textbf{0.96} &  \textbf{0.94} & \textbf{0.62}\\

\bottomrule
\multicolumn{10}{l}{\underline{Underline}: the second best result}
 \end{tabular}
 \caption{Accuracy of all defense baselines under TF, DW and PS attack.}
 \label{tab:results_short}
%  \vspace{-5pt}
\end{table}

\begin{figure}[t!b]
% \vspace{-10pt}
\centerline{\includegraphics[width=0.48\textwidth]{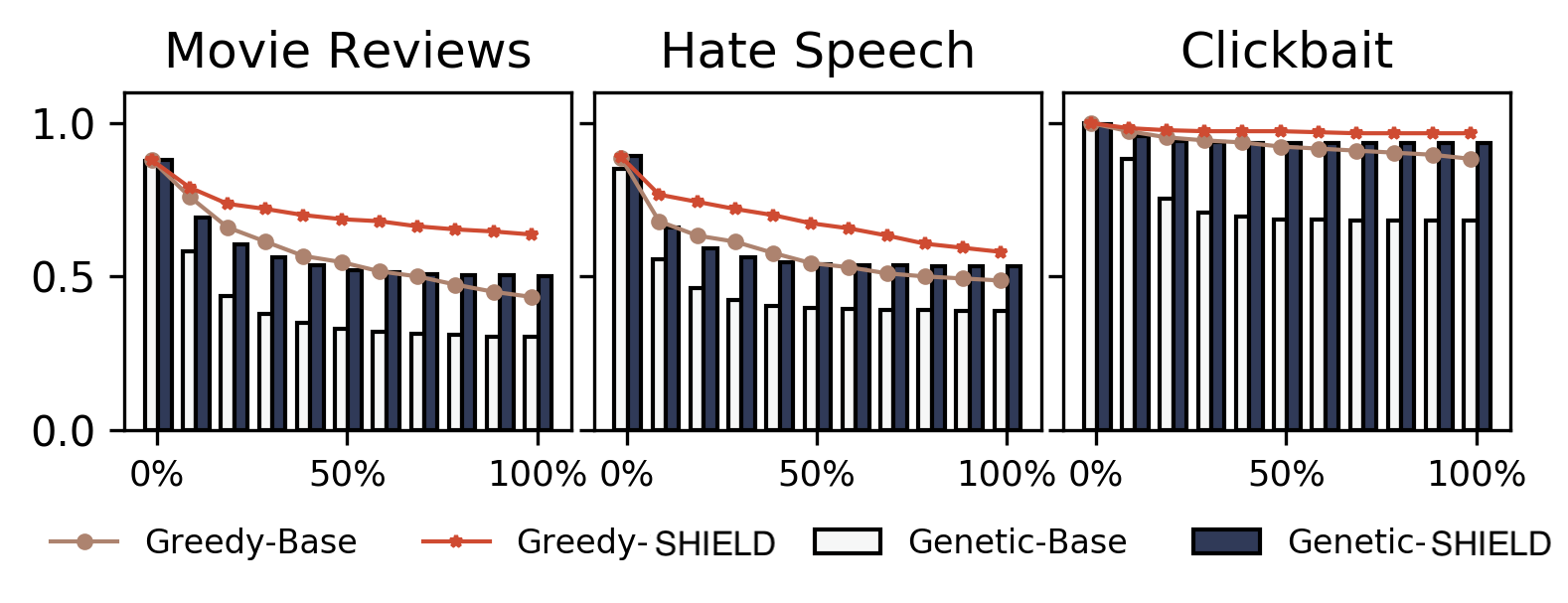}}
\caption{Average accuracy of RoBERTa before and after patched with {\mymethod} under greedy-based and genetic-based attacks with different percentages of \# model queries up to 100\% budget limit.}
\label{fig:numquery}
% \vspace{-15pt}
\end{figure}

% \vspace{0.1in}
\noindent \textbf{Performance under Budgeted Attacks.} {\mymethod}  not only improves the overall robustness of the patched NN model under a variety of black-box attacks, but also induces computational cost that can greatly discourage malicious actors to exercise adversarial attacks in practice. We define computational cost as \# of queries on a target NN model that is required for a successful attack. Since adversaries usually have an attack budget on \# of model queries (e.g. a monetary budget, limited API access to the black-box model), the higher \# of queries required, the less vulnerable a target model is to adversarial threats. A larger budget is crucial for genetic-based attacks because they usually require larger \# of queries than greedy-based strategies. We have demonstrated in Sec. \ref{sec:robustness} that {\mymethod} is robust even when the attack budget is \textit{unlimited}. Fig. \ref{fig:numquery} shows that the performance of RoBERTa after patched by {\mymethod} also reduces at a slower rate compared to the base RoBERTa model when the attack budget increases, especially under greedy-based attacks.

% \begin{table}[t!b]
% \centering
% \footnotesize
% \begin{tabular}{lcccccc}
% \toprule
% \multicolumn{1}{c}{\textbf{Model}} & \multicolumn{3}{c}{\textbf{Train}} & \multicolumn{3}{c}{\textbf{Test}} \\
% \cmidrule(lr){2-4}\cmidrule(lr){5-7}
% {} & \textbf{MR} & \textbf{HS} & \textbf{CB} & \textbf{MR} & \textbf{HS} & \textbf{CB}\\
% \cmidrule(lr){1-7}
% \multirow{1}{*}{CNN+{\mymethod}} & 1e-2 & 1 & 1 & 1e-1 & 1e-1 & 1e-1\\
% \cmidrule(lr){1-7}
% \multirow{1}{*}{RNN+{\mymethod}} & 1 & 1e-2 & 1e-3 & 1e-3 & 1e-3 & 1\\
% \cmidrule(lr){1-7}
% \multirow{1}{*}{BERT+{\mymethod}} & 1e-2 & 1e-1 & 1 & 1e-1 & 1e-2 & 1e-3\\
% \cmidrule(lr){1-7}
% \multirow{1}{*}{RoBERTa+{\mymethod}} & 1 & 1 & 1e-3 & 1e-3 & 1e-3 & 1e-3\\
% \bottomrule
% \end{tabular}
% \caption{Inverse of the final hyper-parameter $\tau$' values for the selected best {\mymethod} model for all datasets.}
% \label{tab:final_tau}
% \end{table}

\vspace{0.1in}
\noindent \textbf{Effects of Stochasticity on {\mymethod}'s Performance}. Stochasticity in {\mymethod} comes from two parts, namely (i) the assignment of the main prediction head during each inference call and (ii) the randomness in the Gumbel-Softmax outputs. Regarding (i), it happens because during a typical iterative black-box process, an attacker tries different manipulations of a given text. When the attacker does so, the input text to the model changes at every iterative step. This then leads to the changes of prediction head assignment because each prediction head is an expert at different features--e.g., words or phrases in an input sentence. Thus, given an input, the assignment of the expert predictors for a specific set of manipulations stays the same. Therefore, even if an attacker repeatedly calls the model with a specific changes on the original sentence, the attacker will not gain any additional information. Regarding (ii), even though Gumbel-Softmax outputs are not deterministic, it always maintains the relative ranking of the expert predictor during each inference call with a sufficiently small value of $\tau$. In other words, it will not affect the fidelity of the model across different runs. 

\vspace{0.1in}
\noindent \textbf{Parameter Sensitivity Analyses.} Training {\mymethod} requires hyper-parameter $K, T, \gamma$ and $\tau$. We observe that arbitrary value $\gamma{=}0.5, K{=}5, T{=}3$ works well across all experiments. 
% A value $\gamma{=}0$ corresponds to {\mymethod} with \textit{only} the SE module, which will be analyzed in Sec. \ref{sec:ablation}. 
Although we did not observe any patterns on the effects of $K$ on the robustness, a $K{\geq}3$ performs well across all attacks. On the contrary, different pairs of the temperature $\tau$ during training and inference witness varied performance w.r.t to different datasets. $\tau$ gives us the flexibility to control the sharpness of the probability vector $\alpha$. When $\tau{\rightarrow}0$, $\alpha$ to get closer to one-hot encoded vector, i.e., use only one head at a time. 
% Table \ref{tab:final_tau} shows the best $\tau$ found using the validation set as explained in Sec. \ref{sec:implementation}.

\vspace{0.1in}
\noindent \textbf{Ablation Tests.}\label{sec:ablation} This section tests {\mymethod} with \textit{only} either the \textit{SE} or \textit{ME} module. Table \ref{tab:ablation} shows that \textit{SE} and \textit{ME} performs differently across different datasets and models. Specifically, we observe that \textit{ME} performs better than the \textit{SE} module in case of Clickbait dataset, \textit{SE} is better than the \textit{ME} module in case of Movie Reviews dataset and we have mixed results in Hate Speech dataset. Nevertheless, the final {\mymethod} model which comprises both the \textit{SE} and \textit{ME} modules consistently performs the best across all cases. This shows that both the \textit{ME} and \textit{SE} modules are complementary to each other and are crucial for {\mymethod}'s robustness.

\section{Limitations and Future Work}
In this paper, we limit the architecture of each expert to be an FCN with a maximum of 3 hidden layers (except the base model). If we include more options for this architecture (e.g., attention~\cite{luong2015effective}), sub-models' diversity will significantly increase. The design of {\mymethod} is model-agnostic and is also applicable to other complex and large-scale NNs such as \textit{transformers-based} models. Especially with the recent adoption of \textit{transformer} architecture in both NLP and computer vision \cite{carion2020end,chen2020generative}, %{\mymethod}, 
potential future work includes extending {\mymethod} to patch other complex NN models (e.g., T5 \cite{raffel2020exploring}) or other tasks and domains such as Q\&A and language generation. Although our work focus is not in robust transferability, it can accommodate so simply by unfreezing the base layers $f(\mathbf{x}, \theta^*_{L-1})$ in Eq. (\ref{eqn:head} during training with some sacrifice on running time. 
% We are also interested in extending our works to a robust transferability setting.
%, and computer vision. 
\renewcommand{\tabcolsep}{1.5pt}
\begin{table}[tb]
\centering
\small
\begin{tabular}{lccccccccc}
\toprule
\textbf{Dataset} & \multicolumn{3}{c}{\textbf{Movie Reviews}} & \multicolumn{3}{c}{\textbf{Hate Speech}} & \multicolumn{3}{c}{\textbf{Clickbait}}\\
\cmidrule(lr){2-4}\cmidrule(lr){5-7}\cmidrule(lr){8-10}
\textbf{Attack} & \textbf{TF} & \textbf{DW} & \textbf{PS} & \textbf{TF} & \textbf{DW} & \textbf{PS} & \textbf{TF} & \textbf{DW} & \textbf{PS} \\

\cmidrule(lr){1-10}
RNN &  0.02 &   0.2 &  0.09 & 0.09 &  0.26 &  0.32 & 0.31 &  0.67 &  0.46 \\
+\textit{SE} Only & \underline{0.02} & \underline{0.17} & \underline{0.08} & 0.09 & \textcolor{black}{0.2} & \underline{0.32} & 0.52 & 0.72 & \underline{0.61} \\
+\textit{ME} Only & \underline{0.02} & 0.14 & 0.07 & \underline{0.13} & \textcolor{black}{0.03} & \textcolor{black}{0.01} & \underline{0.57} & \underline{0.79} & \underline{0.61}\\
\textbf{+\mymethod} &\textbf{0.18} &\textbf{0.44} &\textbf{0.3} &\textbf{0.26} &\textbf{0.61} &  \textbf{0.54} &\textbf{0.78} & \textbf{0.9} &  \textbf{0.85} \\
\cmidrule(lr){1-10}
CNN &  0.01 &  0.13 &  0.06 & 0.03 &   0.1 &  0.14 & 0.45 &   0.7 &  0.57 \\
+\textit{SE} Only & 0.02 & 0.15 & \underline{0.07} & \textbf{0.24} & \textbf{0.42} & \textbf{0.42} & 0.46 & 0.64 & 0.61\\
+\textit{ME} Only & \underline{0.18} & \underline{0.19} & \underline{0.07} & 0.1 & \textcolor{black}{0.25} & 0.29 & \underline{0.60} & \underline{0.80} & \underline{0.69} \\
\textbf{+\mymethod}  & \textbf{0.19} & \textbf{0.38} & \textbf{0.28} & \underline{0.19} & \underline{0.32} & \underline{0.34} &  \textbf{0.74} & \textbf{0.86} & \textbf{0.81}\\
\cmidrule(lr){1-10}
BERT & \underline{0.09} & \underline{0.2} & \underline{0.19} & \underline{0.26} & 0.16 & \underline{0.38} & 0.49 & 0.5 & 0.49\\
+\textit{SE} Only & \textcolor{black}{0.07} & \textcolor{black}{0.18} & \textcolor{black}{0.16} & \underline{0.26} & \underline{0.28} & \textcolor{black}{0.32} & \textcolor{black}{0.45} & \textcolor{black}{0.49} & 0.62\\
+\textit{ME} Only & \textcolor{black}{0.06} & \underline{0.2} & \textcolor{black}{0.15} & \textcolor{black}{0.21} & \underline{0.28} & \textcolor{black}{0.27} & \underline{0.74} & \underline{0.81} & \underline{0.82} \\
\textbf{+\mymethod}  & \textbf{0.26} & \textbf{0.42} & \textbf{0.35} & \textbf{0.37} & \textbf{0.55} & \textbf{0.44} & \textbf{0.92} & \textbf{0.95} & \textbf{0.94}\\

\cmidrule(lr){1-10}
RoBERTa&  0.06 &0.18 &  0.16 &0.1 &0.12 &  0.12 &  0.37 &0.34 &  0.45 \\
+\textit{SE} Only & \underline{0.13} & \underline{0.22} & \underline{0.19} & 0.13 & 0.26 & 0.29 & 0.57 & 0.70 & 0.71\\
+\textit{ME} Only & 0.07 & \textcolor{black}{0.17} & \textcolor{black}{0.15} & \underline{0.22} & \underline{0.4} & \underline{0.31} & \underline{0.8} & \underline{0.87} & \underline{0.85}\\
\textbf{+\mymethod}& \textbf{0.39} &  \textbf{0.55} &  \textbf{0.45} & \textbf{0.37} &  \textbf{0.55} &  \textbf{0.44} & \textbf{0.93} &  \textbf{0.96} &  \textbf{0.94} \\
\bottomrule
% \multicolumn{10}{l}{{\mymethod}\textbackslash ME, {\mymethod}\textbackslash SE: without \textit{ME}, \textit{SE} module.}
\end{tabular}
% \vspace{-10pt}
\caption{Complementary role of \textit{SE} and \textit{ME}.}
\label{tab:ablation}
% \vspace{-10pt}
\end{table}

\section{Related Work}\label{sec:literature}
% \vspace{0.1in}
\noindent \textbf{Defending against Black-Box Attacks.} Most of previous works (e.g., \cite{le2021sweet,zhou2021defense,keller2021bert,pruthi2019combating,dong2021towards,mozes2020frequency,Wang2021AdversarialTW,jia2019certified} in adversarial defense are designed either for a specific type (e.g., word, synonym-substitution as in \textit{certified training}~\cite{jia2019certified}, misspellings~\cite{pruthi2019combating}) or level (e.g., character or word-based) of attack. Thus, they are usually evaluated against a small subset of (${\leq}4$) attack methods. Despite there are works that propose general defense methods, they are often built upon \textit{adversarial training}~\cite{goodfellow2014explaining} which requires training everything from scratch (e.g., \cite{sibetter,miyato2016adversarial,zhang2017mixup} or limited to a set of predefined attacks (e.g., \cite{zhou2019learning}). Although adversarial training-based defense works well against several attacks on BERT and RoBERTa, its performance is far out-weighted by {\mymethod} (Table \ref{tab:results_short}). 

Contrast to previous approaches, {\mymethod} addresses not the characteristics of the resulted perturbations from the attackers but their fundamental attack mechanism, which is most of the time an iterative perturbation optimization process (Fig. \ref{fig:motivation}). This allows {\mymethod} to effectively defend against 14 different black-box attacks (Table \ref{tab:compare_attacks}), showing its effectiveness in practice. To the best of our knowledge, by far, this works also evaluate with the most comprehensive set of attack methods in the adversarial text defense literature.

\vspace{0.1in}
\noindent \textbf{Ensemble-based Defenses.} {\mymethod} is distinguishable from previous ensemble-based defenses on two aspects. First, previous approaches such as DT~\cite{kariyappa2019improving}, ADP~\cite{pang2019improving} are mainly designed for computer vision. Applying these models to the NLP domain faces a practical challenge where training multiple memory-intensive SOTA sub-models such as BERT or RoBERTa can be very costly in terms of space and time complexities. 

In contrast, {\mymethod} enables to ``hot-fix'' a complex NN by replacing and training only the last layer, removing the necessity of re-training the entire model from scratch. Second, previous methods (e.g., DT and ADP) mainly aim to reduce the \textit{dimensionality of adversarial subspace}, i.e., the subspace that contains all adversarial examples, by forcing the adversaries to attack \textit{a single fixed} ensemble of diverse sub-models at the same time. This then helps improve the transferability of robustness on different tasks. However, our approach mainly aims to dilute not transfer but direct attacks by forcing the adversaries to attack stochastic, i.e., different, ensemble variations of sub-models at every inference passes. This helps {\mymethod} achieve a much better defense performance compared to DT and ADP across several attacks (Table \ref{tab:results_short}).

\section{Conclusion}
This paper presents a novel algorithm, {\mymethod}, which consistently improves the robustness of textual NN models under black-box adversarial attacks by modifying and re-training only their last layers. By extending a textual NN model of varying architectures (e.g., CNN, RNN, BERT, RoBERTa) into a stochastic ensemble of multiple experts, {\mymethod} utilizes differently-weighted sets of prediction heads depending on the input. This helps {\mymethod} defend against black-box adversarial attacks by breaking their most fundamental assumption--i.e., target NN models remain unchanged during an attack. {\mymethod} achieves average relative improvements of 15\%--70\% in prediction accuracy under 14 attacks on 3 public NLP datasets, while   still maintaining similar performance on clean examples. Thanks to its model- and domain-agnostic design, we expect {\mymethod} to work properly in other NLP domains.

\section*{Broad Impact Statement}
We address two practical adversarial attack scenarios and how {\mymethod} can help defend against them. First, adversaries can attempt to abuse social media platforms such as Facebook by posting ads or recruitment for human-trafficking, protests, or by spreading misinformation--e.g., vaccine-related. To do so, the adversaries can directly use one of the black-box attacks in the literature to iteratively craft a posting that will not be easily detected and removed by the platforms. In some cases, a good attack method only requires a few trials to successfully fool such platforms. Our method can help confuse the attackers with inconsistent signals, hence reduce the chance they succeed. Second, many popular services and platforms such as the NYTimes, the Southeast Missourian, OpenWeb, Disqus, Reddit, etc. rely on a 3rd party APIs such as \textit{Perspective API}\footnote{\url{https://www.perspectiveapi.com/}} for detecting toxic comments online--e.g., racist, offensive, personal attacks. However, these public APIs have been shown to be vulnerable against black-box attacks in literature~\cite{li2018textbugger}. The attacker can use a black-box attack method to attack these public APIs in an iterative manner, then retrieve the adversarial toxic comments and use those on these platforms without the risk of being detected and removed by the system. Since these malicious behaviors can endanger public safety and undermine the quality of online information, our work has practical values and can have broad societal impacts.

\section*{Acknowledgement}
This research was supported in part by NSF awards \#1820609, \#1915801, and \#2114824. The work of Noseong Park was partially supported by the Yonsei University Research Fund of 2021, and the Institute of Information \& Communications Technology Planning \& Evaluation (IITP) grant funded by the Korean government (MSIT) (No. 2020-0-01361, Artificial Intelligence Graduate School Program (Yonsei University)).

\newpage
\clearpage
\newpage
\bibliography{anthology,custom}
\bibliographystyle{acl_natbib}

\newpage
\clearpage
\newpage
\appendix

\section{ADDITIONAL RESULTS}
\setcounter{table}{0}
\setcounter{figure}{0}
\renewcommand\thetable{\Alph{section}.\arabic{table}}
\renewcommand\thefigure{\Alph{section}.\arabic{figure}}

% \renewcommand{\tabcolsep}{6pt}
% \begin{table}[t!b]
%  \centering
%  \small
%  \begin{tabular}{lcccc}
% \toprule
% \multicolumn{1}{c}{\textbf{Model/Dataset}} & \textbf{MR} & \textbf{HS} & \textbf{CB} & \textbf{AVG}\\
% \cmidrule(lr){1-5}
% CNN   &  0.719 &  \textbf{0.900} &  0.966 &  0.862 \\
% +Ens.  &  0.770 &  0.881 &  \underline{0.975} &  0.875 \\
% +DT        &  0.767 &  0.890 &  0.972 &  0.876 \\
% +ADP       &  0.764 &  0.885 &  \textbf{0.977} &  0.875 \\
% +Mixup     &  0.711 &  0.867 &  0.965 &  0.848 \\
% +AdvT &  \underline{0.772} &  0.884 &  \textbf{0.977} &  \underline{0.878} \\
% +ScRNN     &  0.758 &  0.854 &  0.972 &  0.861 \\
% \textbf{+\mymethod}     &  \textbf{0.787} &  \underline{0.893} &  0.974 &  \textbf{0.885} \\
% \bottomrule
%  \end{tabular}
%  \caption{Prediction performance in F1 on clean examples of CNN-based NN models.}
%  \label{tab:fidelity_full_cnn}
% \end{table}

\renewcommand{\tabcolsep}{5.5pt}
\begin{table}
\centering
\small
\begin{tabular}{lcccccc} 
\toprule
\textbf{Dataset}& \multicolumn{2}{c}{\textbf{MR}}      & \multicolumn{2}{c}{\textbf{HS}}       & \multicolumn{2}{c}{\textbf{CB}}\\
\cmidrule(lr){2-3}\cmidrule(lr){4-5}\cmidrule(lr){6-7}
\textbf{Attack} & \textbf{Bef.} & \textbf{Aft.}& \textbf{Bef.} & \textbf{Aft.} & \textbf{Bef.} & \textbf{Aft.}  \\
\cmidrule(lr){1-7}
SCPNA    & 0.35  & \textbf{0.41}& 0.27  & \textbf{0.4}  & 0.58  & \textcolor{red}{0.53}  \\
TB      & 0.15  & \textbf{0.35}& 0.23  & \textbf{0.48} & 0.79  & \textbf{0.82}  \\
DW     & 0.13  & \textbf{0.38}& 0.1   & \textbf{0.32} & 0.71  & \textbf{0.86}  \\
Kuleshov& 0.01  & \textbf{0.13}& 0.01  & \textbf{0.11} & 0.43  & \textbf{0.63}  \\
TF      & 0.01  & \textbf{0.19}& 0.03  & \textbf{0.19} & 0.44  & \textbf{0.74}  \\
IGA     & 0.05  & \textbf{0.23}& 0.1   & \textbf{0.2}  & 0.6   & \textbf{0.71}  \\
Pruthi  & 0.49  & \textbf{0.54}& 0.47  & \textbf{0.59} & 0.94  & \textcolor{red}{0.9}   \\
PS    & 0.05  & \textbf{0.28}& 0.13  & \textbf{0.34} & 0.56  & \textbf{0.81}  \\
Alzantot & 0.22  & \textbf{0.3} & 0.29  & \textbf{0.36} & 0.82  & \textcolor{red}{0.75}  \\
BAE     & 0.45  & \textbf{0.5} & 0.43  & \textbf{0.55} & 0.77  & \textbf{0.85}  \\
BERTK & 0.0   & \textbf{0.2} & 0.01  & \textbf{0.18} & 0.32  & \textbf{0.61}  \\
PSO     & 0.03  & \textbf{0.03}& 0.23  & \textbf{0.34} & 0.58  & \textcolor{red}{0.56}  \\
Checklist       & 0.7   & \textbf{0.77}& 0.87  & \textbf{0.88} & 0.98  & \textbf{0.98}  \\
Clare   & 0.11  & \textbf{0.3} & 0.48  & \textbf{0.67} & 0.6   & \textbf{0.81}  \\
\cmidrule(lr){1-7}
\textbf{Average} & 0.2   & \textbf{0.33}& 0.26  & \textbf{0.4}  & 0.65  & \textbf{0.75}  \\
\textit{Relative $\uparrow$\%}  & \multicolumn{2}{r}{$\uparrow$65.0\%} & \multicolumn{2}{r}{$\uparrow$53.85\%} & \multicolumn{2}{r}{$\uparrow$15.38\%}  \\ 
\bottomrule
% \multicolumn{7}{l}{\textbf{Bold}, \textcolor{red}{Red}: \textbf{no worse} and \textcolor{red}{decreased} results from the base models}
\end{tabular}
\caption{Accuracy of CNN-based NN models under adversarial attacks before (Bef.) and after (Aft.) being patched with {\mymethod}.}    
\label{tab:results_cnn}
\end{table}

\begin{itemize}[leftmargin=\dimexpr\parindent+0.1\labelwidth\relax]
% \item Table~\ref{tab:fidelity_full_cnn} shows the performance on clean examples of all defense methods on CNN-based NN models.
\item Table~\ref{tab:results_cnn} shows the performance of {\mymethod} against all 14 black-box attacks on CNN-based NN models.
% \item Table~\ref{tab:results_short_cnn} compares the performance of {\mymethod} with all defense baselines on CNN-based NN models. {\mymethod} outperforms all baselines on average.
% \item Table~\ref{tab:ablation_cnn} shows the ablation test of {\mymethod} on CNN-based NN models. 
% \item Table~\ref{tab:final_tau} shows the final $\tau$ parameters found using brute-force search on the validation set as described in Sec. \ref{sec:implementation}. We use this set of parameters to evaluate all the performance under adversarial attacks throughout the paper.
% \item Figure~\ref{fig:k} shows the prediction performance v.s. different number of prediction heads $k$.
\end{itemize}

% \begin{figure}[tb]
% \centerline{\includegraphics[width=0.50\textwidth]{K.png}}
% \caption{Effects of \# of Heads $K$ on the Prediction Performance, $K{=}1$ corresponds to the Base Model.}
% \label{fig:k}
% \end{figure}

% \vspace{-5pt}
\section{REPRODUCIBILITY}\label{sec:reproducibility}
\subsection{Infrastructure and Source Code}
\begin{itemize}[leftmargin=\dimexpr\parindent+0.1\labelwidth\relax]
\item Software: All the implementations are written in Python (v3.7) with Pytorch (v1.5.1), Numpy (v1.19.1), Scikit-learn (v0.21.3). We rely on \textit{Transformers} (v3.0.2) library for loading and training \textit{transformers-based} models (e.g., BERT, RoBERTa).
\item Hardware: We run all of the experiments on standard server machines installed with Ubuntu OS (v18.04), 20-Core Intel(R) Xeon(R) Silver 4114 CPU @ 2.20GHz, 93GB of RAM, and a Titan Xp GPU.
\item Dataset: We use the python library \textit{datasets} (v.1.2.0)~\footnote{\url{https://huggingface.co/docs/datasets/}} by \textit{Hugginface} to load all the benchmark datasets used in the paper.
\item Random Seed: To ensure reproducibility, we set a consistent random seed using \textit{torch.manual\_seed} and \textit{np.random.seed} function for all experiments.
\item Source Code: We will also release the source code of {\mymethod} upon acceptance of this paper.
\end{itemize}

\subsection{Experimental Settings for Base Models}
\subsubsection{\textbf{Architectures and Parameters}}
\begin{itemize}[leftmargin=\dimexpr\parindent+0.1\labelwidth\relax]
\item CNN: We implement the CNN sentence classification model \cite{kim2014convolutional} with three 2D CNN layers, each of which is followed by a \textit{Max-Pooling} layer. Concatenation of outputs of all Max-Pooling layers is fed into a Dropout layer with 0.5 probability, then an FCN + Softmax for prediction. We use an \textit{Embedding} layer of size 300 with pre-trained \textit{GloVe} embedding-matrix to transform each discrete text tokens into continuous input features before feeding them into the CNN network. Each of CNN layers uses 150 kernels with a size of 2, 3, 4, respectively.
\item RNN: Because the original PyTorch implementation of RNN does not support double back-propagation on \textit{CuDNN}, which is required by \textit{DT} and {\mymethod} to run the model on GPU, we use a publicly available \textit{Just-in-Time (JIT)} version of GRU of one hidden layer as RNN cell. We use an \textit{Embedding} layer of size 300 with pre-trained \textit{GloVe} embedding-matrix to transform each discrete text tokens into continuous input features before inputting them into the RNN layer. We flatten out all outputs of the RNN layer, followed by a Dropout layer with 0.5 probability, then an FCN + Softmax for prediction.
\item BERT \& RoBERTa: We use the \textit{transformers} library from \textit{HuggingFace} to fine-tune BERT and RoBERTa model. We use the \textit{bert-base-uncased} version of BERT and the \textit{RoBERTa-base} version of RoBERTa.
\end{itemize}

\subsubsection{\textbf{Vocabulary and Input Length}}
Due to limited GPU memory, we set the maximum length of inputs for transformer-based models, i.e., BERT and RoBERTa, to 128 during training. For CNN and RNN-based models, we use all the vocabulary tokens that can be extracted from the training set, and we use all of the vocabulary tokens provided by pre-trained models for BERT and RoBERTa-based models.

\subsection{Experimental Settings for Defense Methods}
\begin{enumerate}[leftmargin=\dimexpr\parindent+0.1\labelwidth\relax]
\item {\mymethod}: For hyper-parameter $\gamma$, $K$ and $T$, we arbitrarily set $\gamma{\leftarrow}0.5$, $K{\leftarrow}5$ and $T{\leftarrow}3$ and they work well across all datasets. For $\tau$, we already described how to choose the best pair of $\tau$ during training and testing in Sec. \ref{sec:implementation}.
\item \textit{Ensemble}: We train an ensemble model of 5 sub-models, all of which have the same architecture as the base model. We use the average loss of all sub-models as the final loss to train the model.
\item \textit{DT}: We follow the implementation described in Section 3 of the original paper \cite{kariyappa2019improving} and train an ensemble DT model with 5 sub-models, all of which have the same architecture as the base model. We set the hyper-parameter $\lambda\leftarrow 0.5$ as suggested by the original paper.
\item \textit{ADP}: We follow the implementation described in Section 3 of the original paper \cite{pang2019improving} and train an ensemble ADP model with 5 sub-models, all of which have the same architecture as the base model. We set the hyper-parameters required by ADP to default values ($\alpha\leftarrow 1.0$ and $\beta\leftarrow 0.5$) as suggested by the original implementation. 
\item \textit{Mix-up Training (Mix)}: We sample $\lambda \in Beta(1.0,1.0)$ as suggested by the implementation provided by the original paper \cite{zhang2017mixup}.
\item \textit{Adversarial Training}: We use a \textit{1:1} ratio between original training samples and adversarial training samples as suggested by \cite{miyato2016adversarial}. We specifically use the \textit{AT} method as described in Sec. 3 of the original paper~\cite{miyato2016adversarial}.
\item \textit{ScRNN}: We use the implementation and pre-trained model provided by the original paper \cite{pruthi2019combating} that is available at \url{https://github.com/danishpruthi/Adversarial-Misspellings}.
\end{enumerate}

\subsection{Experimental Settings for Attack Methods}
Since we use external open-source \textit{TextAttack}~\cite{morris2020textattack}~\footnote{\url{https://github.com/QData/TextAttack}} and \textit{OpenAttack}~\cite{zeng2020openattack} framework for evaluating the performance of {\mymethod} and all defense baselines under adversarial attacks, implementation of all the attacks are publicly available. Specifically, we use the \textit{TextAttack} framework for evaluating all the word- and character-level attacks, and use the \textit{OpenAttack} for evaluating the sentence-level attack SCPNA. 

\subsection{Experimental Settings for Training and Evaluation}
For every dataset, we train a single {\mymethod} model with the best $\tau$ parameters and evaluate this model with all of the adversarial attacks. In other words, since we have a total of 3 datasets (Movie Reviews, Hate Speech, Clickbait) and 4 base architectures (CNN, RNN, BERT, RoBERTa), we train a total of 12 {\mymethod} models for evaluation. This is done to ensure that we can evaluate the versatility of {\mymethod}'s robustness against different types of attacks \textit{without making any assumptions on their strategies}. During training, we use a \textit{batch size} of 32, \textit{learning rate} of 0.005, \textit{gradient clipping} of 10.0.

For every attack evaluation, we generate a new set of adversarial examples for every pair of \textit{attack method} and \textit{target model}. In other words, since we have a total of 14 different attack methods, 3 datasets, and 4 possible architectures for the base models, this results in a total of 168 different sets of adversarial examples to evaluate in Table~\ref{tab:results}.

\newpage
\clearpage
\newpage
\appendix

\end{document}